\documentclass[10pt,twocolumn,letterpaper]{article}

\usepackage{iccv}
\usepackage{times}
\usepackage{epsfig}
\usepackage{graphicx}
\usepackage{amsmath}
\usepackage{amssymb}

\usepackage{xcolor}
\usepackage{booktabs}
\usepackage{multirow}
\usepackage{graphicx}
\usepackage{threeparttable}
\usepackage{adjustbox}
\usepackage{booktabs}
\usepackage{array}
\usepackage{multicol}
\usepackage[accsupp]{axessibility}
\usepackage[pagebackref=true,breaklinks=true,letterpaper=true,colorlinks,bookmarks=false]{hyperref}

\iccvfinalcopy 

\begin{document}

\title{Attention Where It Matters: Rethinking Visual Document Understanding \\ with Selective Region Concentration}
\author{
   Haoyu Cao\textsuperscript{1\thanks{Equal contribution. \textsuperscript{\dag}Work done during internship at YouTu Lab.}}, \
   Changcun Bao\textsuperscript{1*}, \
   Chaohu Liu\textsuperscript{1,2\dag}, \
   Huang Chen\textsuperscript{1}, \
   Kun Yin\textsuperscript{1}, \
   Hao Liu\textsuperscript{1}, \\
   Yinsong Liu\textsuperscript{1}, \
   Deqiang Jiang\textsuperscript{1}, \
   Xing Sun\textsuperscript{1} \\ 
   Tencent YouTu Lab\textsuperscript{1},  University of Science and Technology of China\textsuperscript{2} \\
   {\tt\small {\{rechycao, changcunbao, huaangchen, zhanyin, ivanhliu, jasonysliu, dqiangjiang}\}@tencent.com} \\
   {\tt\small liuchaohu@mail.ustc.edu.cn, winfred.sun@gmail.com}
}
\maketitle

\ificcvfinal\thispagestyle{empty}\fi
\begin{abstract}
   We propose a novel end-to-end document understanding model called \textbf{SeRum} (SElective Region Understanding Model) for extracting meaningful information from document images, including document analysis, retrieval, and office automation. 
   Unlike state-of-the-art approaches that rely on multi-stage technical schemes and are computationally expensive, 
   SeRum converts document image understanding and recognition tasks into a local decoding process of the visual tokens of interest, using a content-aware token merge module. 
   This mechanism enables the model to pay more attention to regions of interest generated by the query decoder, improving the model's effectiveness and speeding up the decoding speed of the generative scheme. 
   We also designed several pre-training tasks to enhance the understanding and local awareness of the model. 
   Experimental results demonstrate that SeRum achieves state-of-the-art performance on document understanding tasks and competitive results on text spotting tasks.
   SeRum represents a substantial advancement towards enabling efficient and effective end-to-end document understanding.
\end{abstract}

\section{Introduction}
Understanding document images is a fundamental task that involves extracting meaningful information from them, such as document information extraction~\cite{zhang2020trie} or answering visual questions related to the document~\cite{docvqa}.
In today's world, where the volume of digital documents is increasing exponentially, this task has become even more critical in various applications, including document analysis~\cite{cui2021document}, document retrieval~\cite{mass2020docret}, and office robotic process automation (RPA)~\cite{axman2020rpa}.

\begin{figure}[t]
   \centering
   \includegraphics[width=0.5\textwidth]{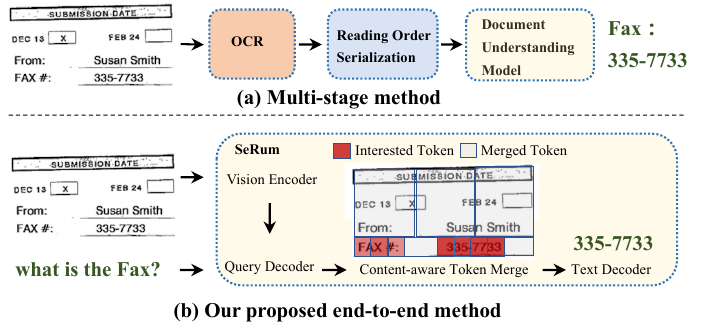}
   \caption{Comparison of multi-stage and SeRum end-to-end processing flows for visual document understanding. SeRum's approach simplifies the pipeline by directly generating text output for \textit{key visual tokens} from the image, eliminating the need for OCR and delivering highly efficient and effective document analysis. Best viewed in color.}
   \label{fig:motivation}
   \vspace{-0.1cm}
\end{figure}

Current state-of-the-art approaches~\cite{huang2022layoutlmv3} rely on multi-stage technical schemes involving optical character recognition (OCR)~\cite{van2020ocr} and other modules~\cite{qian2018graphie} to extract key information, as shown in Figure~\ref{fig:motivation}. 
However, these approaches are suboptimal and computationally expensive, relying too much on prefacing modules such as accurate OCR recognition and document content ordering~\cite{hwang2019post}.

To address these challenges, we propose a novel end-to-end document understanding model with selective region concentration called \textbf{SeRum} (SElective Region Understanding Model). 
As shown in Figure~\ref{fig:motivation}, SeRum converts document image understanding and recognition tasks into a local decoding process of the visual tokens of interest, which includes a vision encoder, a query-text decoder, and a content-aware token merge module.

For document understanding tasks, the content to be extracted often takes up a small proportion of the whole document area but may change greatly in scale. 
Therefore, it is crucial to accurately identify the key area of interest first. 
SeRum extracts document image features using a vision Transformer-based encoder. 
We use a self-encoding Transform query decoder inspired by MaskFormer~\cite{maskformer}, which decode the input query (question for tasks) and carries out cross-attention mechanism with image features to form the embeddings of queries. 
Then, we obtain the area of interest mask through dot product with the up-sampled image features. 
Since the number of queries is greater than the number of text locations required, we use binary matching for pairing, following DETR~\cite{detr}.

The final sequence output is automatically generated by the text decoder through cross-attention with the encoded visual token. 
However, the presence of noise in long visual token sequence could adversely affect the decoding process. 
To address this issue, we propose a content-aware token merge mechanism that selects visual tokens associated with the query while merges the rest. 
This mechanism constrains attention to regions of interest generated by the query decoder, while simultaneously preserving global information and enhancing regional information of interest. 
The multi-query mechanism employed in our approach enables local generation of text, thereby resulting in shorter and more precise text content.

To further enhance the understanding and local awareness of the model, we design three pre-training tasks, including query to segmentation, text to segmentation and  segmentation to text. 
In summary, we propose a novel end-to-end document understanding model called SeRum that improves the recognition ability of end-to-end models while achieving competitive results in word recognition. 
Our content-aware token merge mechanism limits the decoder's attention to the local details of interest, improving the model's effect and speeding up the decoding speed of the generative scheme. 
We believe that the SeRum model offers a valuable step towards efficient and effective end-to-end document understanding, with potential applications in various fields such as automatic document analysis, information extraction, text recognition and \textit{etc.}
Conclusively, our contributions are summarized into the three folds:
\begin{itemize}
   \item We propose a novel end-to-end document understanding model called SeRum, which converts document image understanding and recognition tasks into a local decoding process of the interested visual tokens.
   \item We introduce a content-aware token merge mechanism that improves the model's perception of image details and speeds up the decoding speed of the generative scheme. 
   \item Experimental results on multiple public datasets show that our approach achieves state-of-the-art performance on document understanding tasks and competitive results on text spotting tasks.
\end{itemize}

\section{Related Work}
Document image understanding has been a significant topic in computer vision for many years and has attracted considerable attention from the research community~\cite{cui2021document, docrl}. 
Traditional rule-based methods~\cite{ha1995recursive, simon1997fast} and machine learning-based techniques~\cite{marinai2005artificial, shilman2005learning} often involve manual feature selection, which can be insufficient when dealing with complex layout documents. 
With the advent of deep learning technology, document information extraction methods have witnessed notable improvements in both effectiveness and robustness.
In recent years, deep learning-based methods for document understanding can be categorized into three primary groups: OCR-dependent method with post-processing, OCR-dependent method without post-processing, and end-to-end method without OCR.

\noindent\textbf{OCR-dependent methods with post-processing} involve extracting textual information from document images using OCR engines and other auxiliary modules. 
In general, the OCR engine first detects and recognizes the text content of the document, then sorts the text according to the reading order, and finally marks each words in the way of sequence annotation through the document understanding model. 
In this way, the document understanding model focuses on the representation of document content and numerous significant works have arisen in this field. 
One such example is the LayoutLM family~\cite{xu2020layoutlm, xu2020layoutlmv2, xu2021layoutxlm, huang2022layoutlmv3}, which employs multimodal networks to integrate visual and text features. 
GCN (Graph Convolution Network) is another technique utilized to model the intricate relationship between text and image~\cite{qian2018graphie, liu2019graph, wei2020robust, yu2021pick, lockard2019openceres, lockard2020zeroshotceres}. Additionally, Chargrid~\cite{katti2018chargrid, denk2019bertgrid} leverages layout information to extract entity relationships.

\noindent\textbf{OCR-dependent methods without post-processing} are designed to address the issue of token serialization errors in post-processing, and to mitigate OCR errors to a certain extent. 
For example, GMN~\cite{cao2022gmn} employs an optimally structured spatial encoder and a modality-attentive masking module to tackle documents of intricate structure which are hard to serialize. 
QGN~\cite{cao2022qgn} utilizes a query-based generation scheme to bolster the generation process even with OCR noise. 
Other works such as~\cite{sage2020pgn, powalski2021t5, tang2022udop} encode the derived document information as a succession of tokens in XML format and output the XML tags delimiting the types of information.

\noindent\textbf{End-to-end methods without OCR} aim to further eliminate the dependence on the OCR recognition module, which exhibits faster reasoning speed and necessitates fewer parameters. 
These methods have recently gained attention due to their ability to achieve higher efficiency and effectiveness. 
For instance, Donut~\cite{kim2022donut} and Dessurt~\cite{davis2023dessurt} employ Swin Transformer~\cite{liu2021swin} as an encoder for extracting image features, and BART-like Transformers~\cite{lewis2019bart} as a decoder for generating text sequences.

\begin{figure*}[t]
   \centering
   \includegraphics[width=\linewidth]{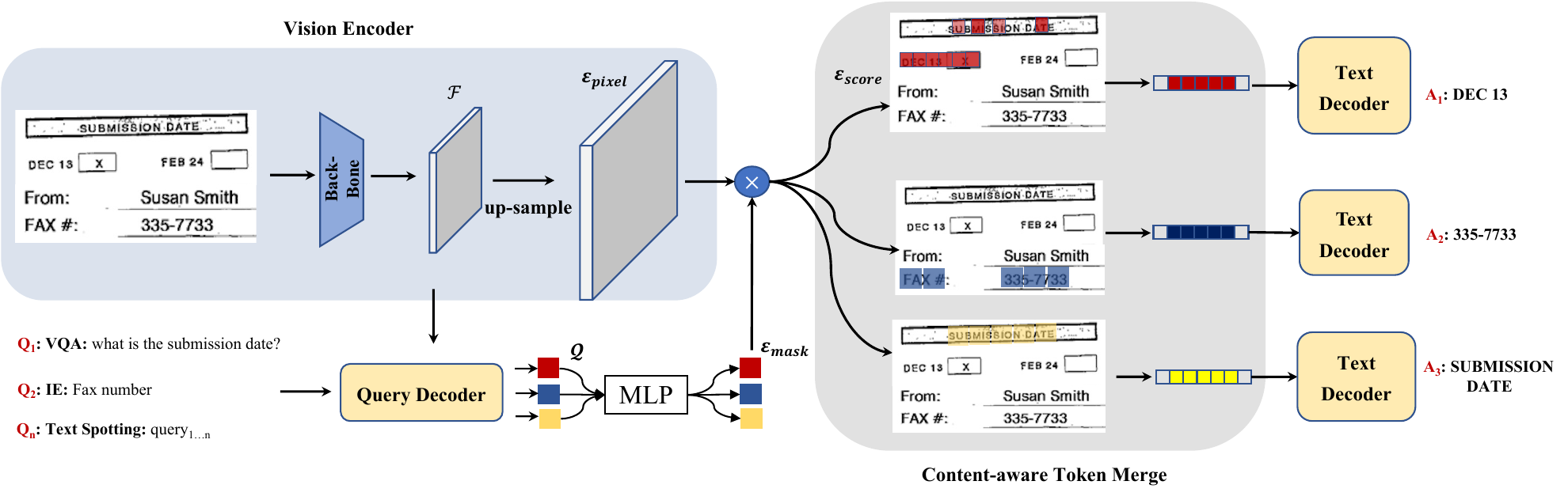}
   \caption{The architecture of our proposed SeRum. Given an input image and query, the vision encoder and query decoder encode the image and extract the region mask of interest, respectively. 
   The resulting mask is then passed through the content-aware token merge module, which filters out the Top-K visual tokens of interest and merges the remaining ones. Finally, the resulting merged tokens are passed through the text decoder to obtain the document understanding result. Best viewed in color.}
   \label{fig:overall}
   \vspace{-0.1cm}
\end{figure*}

While the above solutions have addressed the document understanding dependency on OCR, they are all full graph encoding and global decoding. 
As such, they may not be suitable for scenarios where OCR identification errors occur due to the inability to pay attention to the local details of the document, leading to low generation efficiency. 
The SeRum model proposed in this work offers a valuable step towards efficient and effective end-to-end document understanding and serves as a bridge for OCR recognition and document understanding.

\section{Method}
\subsection{Overall Architecture}
SeRum is an end-to-end architecture designed to tackle document understanding tasks by leveraging a process of decoding visual regions of interest. 
As shown in Figure~\ref{fig:overall}, it consists of three main components: vision encoder, query-text decoder, and content-aware token merge modules.
The vision encoder employs a visual transformers backbone to extract image features $\mathcal{F}$ which are then upsampled to form high-resolution feature embeddings $\mathcal{E}_{\text {pixel }}$.
The query decoder is a transformer decoder that encodes the input queries and attends to the image features. 
It produces $N$ per-segment embeddings $\mathcal{Q}$ that independently generate $N$ class predictions with $N$ corresponding mask embeddings $\mathcal{E}_{\text {mask }}$.
The model then predicts $N$ possibly overlapping binary mask $\mathcal{E}_{\text {score }}$ predictions via a dot product between vision embeddings $\mathcal{E}_{\text {pixel }}$ and mask embeddings $\mathcal{E}_{\text {mask }}$, followed by a sigmoid activation. 
The content-aware token merge module merges irrelevant tokens, and via a text decoder, generates the final text token result.
By using a transformer decoder to encode the input query, the model can effectively reason about the document's semantic content and produce informative predictions. 
The content-aware token merge modules leverage the model's attention to combine relevant tokens and improve the accuracy of the final predictions. 

\subsection{Vision Encoder}
The vision encoder module plays a crucial role in converting the input document image $\mathbf{x} \in \mathbb{R}^{H \times W \times C}$ 
into a feature map $\mathcal{F} \in \mathbb{R}^{h \times w \times d}$ that can be processed by subsequent modules.
The feature map is further serialized into a set of embeddings 
$\left\{\mathbf{z}_i \mid \mathbf{z}_i \in \mathbb{R}^d, 1 \leq i \leq n\right\}$
where $n$ represents the feature map size or the number of image patches, and $d$ is the dimension of the latent vectors of the encoder.
CNN-based models~\cite{resnet} or Transformer-based models such as the Swin Transformer~\cite{liu2021swin} can be used as the encoder network. 
For this study, we utilize the Swin Transformer with modifications as it exhibits superior performance in our preliminary study on document parsing. 
The Swin Transformer divides the input image $\mathbf{x}$ into non-overlapping patches, and applies Swin Transformer blocks consisting of a shifted window-based multi-head self-attention module and a two-layer MLP to these patches. 
Patch merging layers are subsequently applied to the patch tokens at each stage, and the output of the final Swin Transformer block ${\mathbf{z}}$ is fed into the query-text decoder. We also upsample the feature map to a larger size,  $\mathcal{E}_{\text {pixel }} \in \mathbb{R}^{sh \times sw \times d}$ for fine-grained mask area acquisition, where $s$ is the upsampled factor and we set $s=6$ as default.
In addition to the Swin Transformer architecture, we introduce a learnable position embedding in the last layer of the vision encoder module, which enhances the model's perception of location. 
The position embedding is added to the upsampled feature map:
$ \mathcal{E}_{\text{pixel}}^{'}=\mathcal{E}_{\text{pixel}}+\mathcal{P} $,
where $\mathcal{E}_{\text{pixel}}^{'}$ is the upsampled feature map with the position embedding, $\mathcal{E}_{\text{pixel}}$ is the original upsampled feature map, and $\mathcal{P}$ is the learnable position embedding.
\subsection{Query-Text Decoder}
The Query-Text decoder is an essential module in the field of visual contextual analysis. It comprises two sub-modules, namely the query decoder and text decoder, which operate with shared weights.

\textbf{Query Decoder} utilizes the standard Transformer~\cite{transformer} to compute the per-segment embeddings $\mathcal{Q} \in \mathbb{R}^{C_{\mathcal{Q}} \times N}$, $\mathcal{E}_{\text {mask }} \in \mathbb{R}^{N \times d}$ encoding global information about each segment prediction from the image features $\mathcal{F}$ and $N$ learnable token embeddings or questions.
Following the approach of \cite{maskformer}, the decoder produces all predictions simultaneously. 

\textbf{Text Decoder} is designed to decode the corresponding text token sequences for a variety of tasks, including Visual Question Answering (VQA), information extraction and text spotting. It generates text in an auto-regressive manner by attending to the query embedding, previously generated tokens, and the encoded visual features. The mathematical form of the text decoder is as follows:
\begin{equation}
\mathbf{h}_t=\operatorname{Decoder}\left(\mathbf{h}_{t-1}, \mathcal{F}, \mathcal{Q}, \mathcal{T}_{t-1}\right)
\end{equation}
Here, $\mathbf{h}_t$ represents the hidden state of the decoder at time step $t$, $\mathcal{Q}$ represents the query embedding, $\mathcal{T}_{t-1}$ represents the previous token embedding, and $\mathcal{F}$ represents the visual feature map. In contrast, the query decoder is a self-encoding form, where each query attends to all other queries. The mathematical form is as follows:
\begin{equation}
   \mathbf{h}_i=\operatorname{Decoder}\left(\mathcal{F}, \mathcal{Q}_i,{\mathcal{Q}_j \mid j \neq i}\right)
\end{equation}
Here, $\mathbf{h}_i$ represents the hidden state of the decoder for query $i$, and ${\mathcal{Q}_j \mid j \neq i}$ is the set of all other queries.

\subsection{Content-aware Token Merge}
Document visual understanding tasks differ from conventional visual tasks in that a amount of information is usually concentrated in a very small area, commonly comprising less than 5\% of the entire image. 
Therefore, the visual tokens extracted by the vision module contain a large number of background areas, which increase the difficulty of decoding and reduce the decoding speed. 
To address this issue, we introduce a module called content-aware token merge that dynamically focuses on the more relevant foreground part of the visual tokens and reduces the focus on the background part of the tokens.

\textbf{Foreground Area}: As mentioned above, the correlation between query and visual tokens can be represented by $\mathcal{E}_{\text {score }} \in \mathbb{R}^{N \times sh \times sw}$. 
The higher the score, the stronger the correlation information of the visual token representing the location. 
We sort the $\mathcal{E}_{\text {score }}$ and select the features corresponding to the top-K tokens with high scores to form a \textit{foreground area sequence} for each query. 
$\mathcal{E}_{\text {score}}^{l}$ denotes the average score associated with token $l$.
\begin{equation}
\mathbb{F}_f=\left[\mathbf{z}_l * \mathcal{E}_{\text {score}}^{l} \mid l=1, \ldots, K\right]
\end{equation}

\textbf{Background Area}: 
The foreground area features represent the visual features of absolute dominance, 
while the rest of the features are considered background area features. 
These features are typically global features or features used to assist in extrapolation. 
To preserve this part of the features and avoid interference with the foreground features caused by excessive features of this part, 
In our case, we project the background tokens using the basic attention mechanism onto the foreground tokens, resulting in a K-dimensional feature representation that captures the important information from both the foreground and background tokens.
The formal formula for the attention mechanism is as follows:
\begin{equation}
\mathbb{F}_b = \text{softmax} (\mathbf{Q} \mathbf{f}_{r}^\intercal) (\mathbf{W}^v \mathbf{f}_{r}^\intercal) ^ \intercal  
\end{equation}
where $\mathbb{F}_b \in \mathbb{R}^{K \times d}$ are merged background features, $\mathbf{Q} \in \mathbb{R}^{K \times d}$ are foreground query features, $\mathbf{f}_{r} \in \mathbb{R}^{(L-K) \times d}$ are primitive background features, and $\mathbf{W}^v \in \mathbb{R}^{d \times d}$ is the projection weight. 
The softmax function is employed to normalize the dot product of the query and key matrices.
Subsequently, the resulting weights are applied to the value matrix to derive the output.
This process yields a new feature vector $\mathbb{F}_{\text{b}}$, which encapsulates the background contextual information. 
The visual context is represented by $\mathbb{F} = \mathbb{F}_f +  \mathbb{F}_b$.

To adapt to varying size changes, we define the number of foreground tokens, denoted by $K$, as a function of the total number of tokens, represented by $L$, i.e., $K=\alpha L$. 
During training, we sample $\alpha$ from a uniform distribution that ranges from 0.02 to 1.0. 
Specifically, $\alpha=1.0$ indicates that no token merge occurs. During inference, the value of $\alpha$ can be fixed based on the performance requirements.

\subsection{Pre-training\label{sec:pretraining}}
In this study, we utilize multi-task pre-training to enhance the model's position understanding and text generation capabilities. 
The pre-training process involves three subtasks: query to segmentation, text to segmentation, and segmentation to text.

\textbf{Query to Segmentation} aims to equip the model with text detection capability. 
We adopt the query generation approach used in DETR~\cite{detr} but apply it to an instance segmentation task instead, as text generation is a highly dense prediction task. 
We utilize the vision encoder and query decoder to process a given image $\mathbf{x}$ with $N$ learnable token embeddings, resulting in $N$ mask predictions $\mathcal{E}_{\text {score }}$ for the text area,
and we set $N=50$ as the default.

\textbf{Text to Segmentation} helps the model to comprehend the distinct positions of each text element, thus playing a crucial role in text-segmentation alignment. 
We feed the text snippet of the image to query decoder and turn this into an instance segmentation task as well. 
Since document images often contain a relatively low proportion of pixels with text, emphasizing the text region segmentation performance can enhance the quality of the decoder generated output.

\textbf{Segmentation to Text} aims to generate text from image segmentations in a manner similar to OCR. 
During this stage, the foreground and background features $\mathbb{F}$ generated from the preceding layers is utilized as the visual context within the cross-attention mechanism of the text decoder, which auto-regressively generates tokens.

It is important to note that the three pre-training subtasks are performed simultaneously and with no particular hierarchical order. 
In other words, the training process includes data from all three subtasks in the same batch.

\subsection{Training Strategy}

\textbf{Loss Function}.
SeRum is trained end-to-end with a loss function consisting of multiple parts. 
The first part is the Hungarian matching loss between queries and targets, which is used to handle the case where the number of queries is greater than the number of targets. 
Specifically, we conduct position matching at each layer of the query encoder and position and category matching at the last layer. 
The loss for this part is defined as:

\begin{equation}
\mathcal{L}_{match} = \sum_{i=1}^{N_e} L_{match}^{(i)}
\end{equation}
where $N_e$ is the number of query encoder layers, and $L_{match}^{(i)}$ is the matching loss at layer $i$.

The second part of the loss is the autoregressive decoder loss for the text decoder, which is defined as:
\begin{equation}
\mathcal{L}_{decoder} = -\log P(y_{1:T}|\mathbf{h}_{1:T}, \mathbf{e}_1)
\end{equation}
where $y_{1:T}$ is the ground-truth text, $\mathbf{h}_{1:T}$ is the hidden states of the decoder, and $\mathbf{e}_1$ is the embedding of the start-of-sequence token.

To speed up the learning of areas of interest, we add a text constraint loss to the features of the image up-sampled above, where the mask of the loss constraint is changed to cover all text areas in the whole image. The loss for this part is defined as:
\begin{equation}
\mathcal{L}_{text} = \frac{1}{|\Omega|}\sum_{(i,j)\in\Omega} L_{text}(\mathcal{E}^{i,j}_{\text {score }})
\end{equation}
where $\Omega$ is the set of all text areas in the image, $L_{text}$ is the text constraint loss, and $\mathcal{E}^{i,j}_{\text {score }}$ is the segmentation mask at position $(i,j)$.

The total loss is a weighted sum of the above loss parts:
\begin{equation}
\mathcal{L}_{total} = \lambda_{1}\mathcal{L}_{match} + \lambda_{2}\mathcal{L}_{decoder} + \lambda_{3}\mathcal{L}_{text}
\end{equation}
where $\lambda_{1}$, $\lambda_{2}$, and $\lambda_{3}$ are hyper-parameters controlling the weights of each loss part.

\textbf{More Details}.
During pre-training, SeRum employs a binary matching mode between query and target and has explicit supervision in the area of interest mask. However, during downstream task training, the model is constrained by the proportion of the area of interest, leading to an implicit supervision mode.

\section{Experiments}
We evaluate the proposed model on three tasks: document information extraction, document visual question answering, and text spotting. 
To assess the model's performance, we compare it with state-of-the-art two-stage document information extraction methods as well as other end-to-end methods.

\subsection{Tasks and Datasets}

\noindent\textbf{Document Information Extraction (DIE)} is a process of extracting key-value pairs structured data from documents. 
We evaluate our model's performance on three benchmark datasets commonly used to validate the efficiency of the models in DIE.

\noindent\textbf{\textit{Ticket}} dataset \cite{guo2019eaten} includes 300,000 composite images and 1,900 real images of Chinese train tickets. 
There are eight entities, such as start station, terminal station, time, price, \textit{etc.,} that need to be extracted. 

\noindent\textbf{\textit{CORD}} dataset \cite{park2019cord} is a widely-used English benchmark for information extraction tasks. 
It comprises 800 training, 100 validation, and 100 test receipt images, with 30 distinct subclasses for extraction, such as menu name, menu num, menu price, \textit{etc.} 
CORD is complex due to its multi-layered, nested structure, which requires the model not only to extract the relevant text but also to gain a deeper understanding of its structure.

\noindent\textbf{\textit{SROIE}} dataset \cite{huang2019icdar2019} consists of 973 scanned receipt images, divided into 626 for training and 347 for testing purposes. 
Its composition is simplistic, encompassing solely four key entities: company, total, date, and address. 
Nevertheless, due to the dense textual content present in the images, extracting relevant information is complicated, as some entities are spread across multiple lines.

We utilize two common metrics, F1 score~\cite{hong2022bros, xu2020layoutlmv2} and Tree Edit Distance (TED)~\cite{kim2022donut, edit}, to demonstrate the model's performance on the datasets. F1 score is a harmonic mean of precision and recall of a classification model, where precision measures the ability of the model to correctly identify positive cases, and recall measures the ability of the model to identify all positive cases. A high F1 score indicates that the model has high accuracy and precision in classifying positive cases. Although F1 score is very intuitive, it is relatively strict and cannot accurately reflect the prediction accuracy at the character level. The TED is the minimum number of single-character editing operations required to convert one string to another. We calculate its score through $max(0, 1 - TED(pr, gt) / TED(\phi, gt))$, where $gt$, $pr$, and $\phi$ stand for ground truth, predicted, and empty string.

\noindent\textbf{DocVQA} is a task that combines document understanding and visual question answering. The goal is to answer a question about a given document image. We evaluate our model on the DocVQA dataset \cite{docvqa}, which contains 12,767 document images and 99,000 questions in total. The dataset is split into training, validation, and test sets, with 80\%, 10\%, and 10\% of the data in each set, respectively. The questions are of various types, such as yes/no, counting, reasoning, and comparison, and require both text understanding and visual reasoning. The ANLS (Average Normalized Levenshtein Similarity) metric, which is an edit-distance-based metric, is used for evaluation.

\subsection{Implementation Details}
\textbf{Model Architecture.} 
Our proposed model, SeRum, utilizes Swin-B~\cite{liu2021swin} as the visual encoder with slight modifications. 
Specifically, we set the layer numbers and window size as {2, 2, 14, 2} and 10 to extract features from the input image. 
Moreover, the input image resolution is set to $1280 \times 960$.
To balance the trade-off between speed and accuracy, we use the first four layers of mBART as the decoder to generate the output text.

\textbf{Pre-training.}
To improve the model's performance, we pretrain SeRum on a large-scale synthetic dataset generated using the Synthetic Text Recognition (SynthText) \cite{synthtext}, Synth90K~\cite{Synth90K}, IIT-CDIP~\cite{iit-cdip} datasets and multi-language synthetic dataset following Donut. 
The SynthText dataset is used to generate text instances with complex backgrounds, while the Synth90K dataset is used for text instances with simple backgrounds. 
The IIT-CDIP dataset comprises over 11M scanned images of English language documents.
We use the multi-task pre-training strategy, as described in Section~\ref{sec:pretraining}.

\textbf{Fine-tuning.}
For the fine-tuning stage, we utilize the Adam optimizer with an initial learning rate of $5\times10^{-5}$ for all datasets, and incorporate a learning rate decay factor of 0.1 after every 30 epochs to enhance the model's performance. 
For the information extraction tasks, we set the batch size to 24, while for the DocVQA task, we set it to 8. 
To ensure optimal performance and convergence, we train the model for a maximum of 300 epochs. 
Additionally, we employ two generation mechanisms, SeRum-total and SeRum-prompt, for information extraction tasks. SeRum-total generates a complete token sequence of all key information using a predefined format, such as Donut, with the task name as queries. 
SeRum-Prompt uses the keys as queries and generates each information in parallel. We set the token keep ratio $\alpha$ to 0.5 for SeRum-total generation mechanism and $\alpha$ to 0.1 for SeRum-Prompt generation mechanism during test stage. 
Finally, to balance the loss function weights, we set $\lambda_{1} = \lambda_{2} = \lambda_{3}$.

\subsection{Comparisons with Previous Approaches}
\begin{table*}[t]
   \centering
   \begin{adjustbox}{max width=\textwidth}
   \begin{threeparttable}
     \centering
     \begin{tabular}{lcccccccc|cc}
     \toprule
      & & & \multicolumn{2}{c}{CORD~\cite{park2019cord}} & \multicolumn{2}{c}{Ticket~\cite{guo2019eaten}} & \multicolumn{2}{c}{SROIE~\cite{huang2019icdar2019}} & \multicolumn{2}{|c}{DocVQA~\cite{docvqa}}\\
      \cmidrule(lr){4-5} \cmidrule(lr){6-7} \cmidrule(lr){8-9} \cmidrule(lr){10-11}
       & OCR & \#Params & F1 & Acc. & F1 & Acc. & F1 & Acc. & ANLS & ANLS$^{\ast}$ \\
       \midrule
       SPADE~\cite{hwang2020spade}  & \checkmark & $93\text{M} + \alpha^{\ddag}$ & 74.0 & 75.8 & 14.9 & 29.4 & - & - & - & - \\
       WYVERN~\cite{hwang2020towards}  & \checkmark & $106\text{M} + \alpha^{\ddag}$ & 43.3 & 46.9 & 41.8 & 54.8 & - & - & - & - \\
       BERT~\cite{hwang2019post} & \checkmark & $86{\text{M}}+\alpha^{\ddag}$ & 73.0 & 65.5 & 74.3 & 82.4 & - & - & 63.5 & - \\
       LayoutLMv2~\cite{xu2020layoutlmv2} &\checkmark& $179{\text{M}}+\alpha^{\ddag}$ & 78.9 & 82.4 & 87.2 & 90.1 & 61.0 & 91.1 & \textbf{78.1} & 67.3 \\
       \midrule
       Donut~\cite{kim2022donut} & & $143{\text{M}}$ & 84.1 & 90.9 &94.1 & 98.7 & 83.2 & 92.8 & 67.5 & 72.1\\
       \textbf{SeRum-total} & & $136{\text{M}}$ & 80.5 & 85.8 & 97.9 & 99.6 & 85.6 & 92.8 & - & -\\
       \textbf{SeRum-prompt} & & $136{\text{M}}$ & \textbf{84.9} & \textbf{91.5} & \textbf{99.2} & \textbf{99.8} & \textbf{85.8} & \textbf{95.4} & 71.9 & \textbf{77.9} \\
       \bottomrule
     \end{tabular}

   \end{threeparttable}
   \end{adjustbox}
   \caption{{\bf Performances on various document understanding tasks.} The field-level F1 scores and tree-edit-distance-based accuracies are reported. \textbf{SeRum-prompt} shows competitive in different tasks. Parameters for vocabulary are omitted for fair comparisons among multi-lingual models. $^\ddag$\# parameters represents that OCR needs to be used. }
    \label{table:results}
\end{table*}

\noindent\textbf{Document Information Extraction.}
We present a comparative evaluation of our proposed method with several state-of-the-art approaches reported in recent years on three widely-used datasets, namely Ticket, CORD, and SROIE.
Our comparison includes approaches that employ OCR as well as fully end-to-end methods.
In the former, OCR is used to extract the text and position of the image, and the information is sorted and classified at the token level, after which the target information is determined based on the classification results.
Approaches such as BERT~\cite{hwang2019post}, LayoutLMv2~\cite{xu2020layoutlmv2}, and SPADE~\cite{hwang2020spade} can produce satisfactory results when the OCR results are entirely accurate.
We use the OCR engine API reported in Donut, including MS OCR and others, for consistency.
In addition, we evaluate a generative method called WYVERN~\cite{hwang2020towards} that utilizes the encoder and decoder architecture of Transformer and requires OCR.
End-to-end methods, which have a streamlined pipeline, are increasingly popular in both industry and academia.
Donut~\cite{kim2022donut} and Dessurt~\cite{davis2023dessurt} are two prominent end-to-end methods, and we only study the results of Donut in this paper due to the underperformance of Dessurt.

\begin{figure*}[!htbp]
   \centering
   \includegraphics[width=1.0\textwidth]{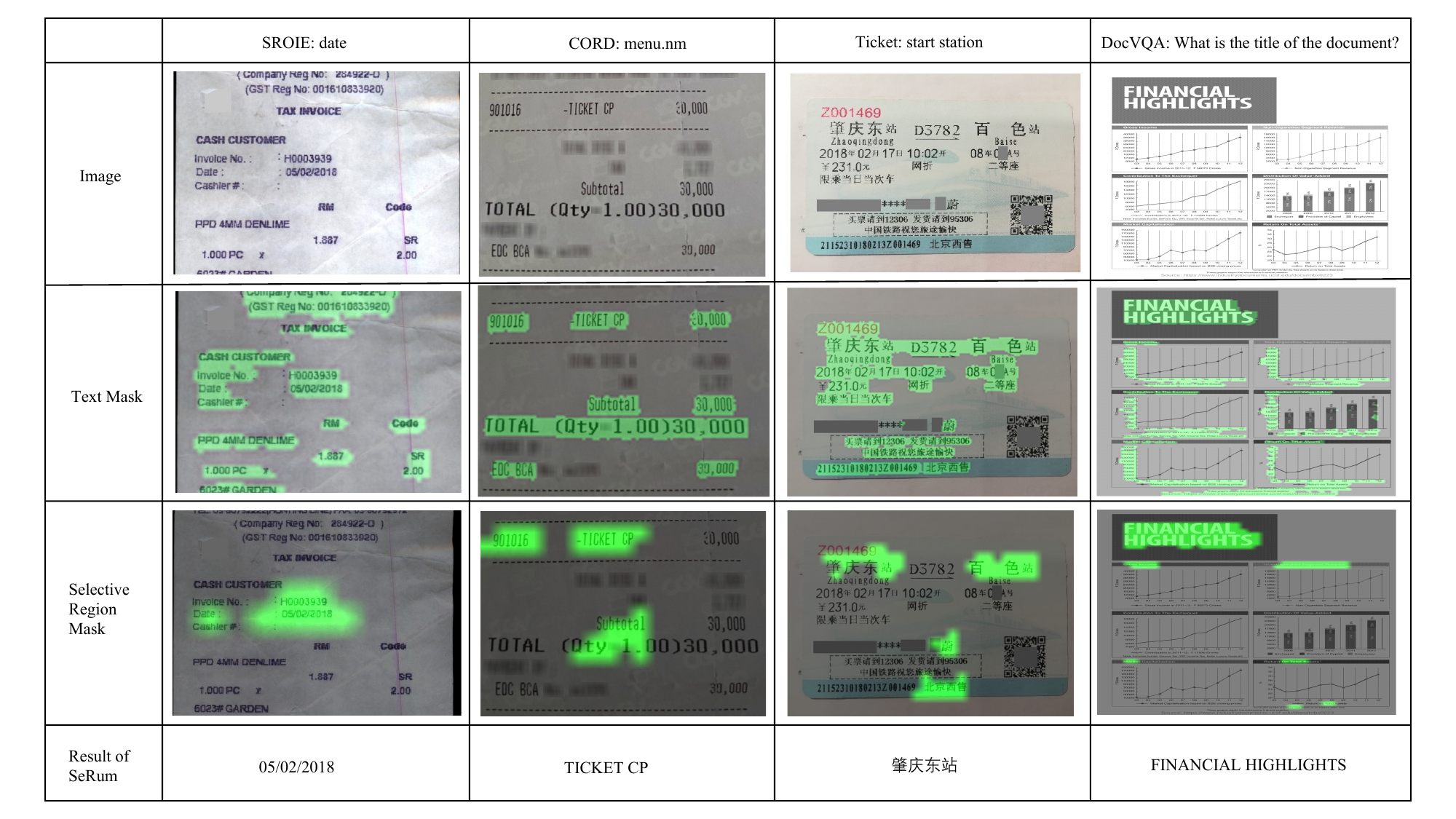}
   \caption{Visualization of SeRum on various datasets. 
   The figure depicts five row from top to bottom, showcasing the task and query, 
   original image, text regions of pre-training model, selected region of interest obtained by the query, and corresponding text result SeRum extracted.
   The results showed that SeRum effectively captured the regions of interest and decoded them correctly. Best viewed in color.
   }
   \label{fig:showcase}
\end{figure*}

Our proposed method achieves new state-of-the-art results on the three open benchmarks of document information extraction, as shown in Table~\ref{table:results}.
Our model excels in the Ticket dataset, achieving a score of over 99\% and demonstrating near-complete success in addressing this task, outperforming the second best end-to-end method Donut by 5\%.
Moreover, our model exhibits robust character recognition and context understanding capabilities, as shown in Figure \ref{fig:showcase}.
Notably, SeRum is an end-to-end method that does not require an OCR module, making it more efficient for training and inference, and therefore suitable for industrial applications.
In particular, on the SROIE dataset, the F1 score of SeRum exceeds that of the multi-stage method LayoutLMv2 by 24\%.

\noindent\textbf{DocVQA.} 
We conduct an evaluation of our proposed SeRum model on the challenging DocVQA dataset and compare its performance against several state-of-the-art methods. 
The experimental results are summarized in Table \ref{table:results}. 
Our SeRum model demonstrates superior performance over the strong baseline of end-to-end methods and achieves competitive results with the multi-stage methods. 
In particular, in the ANLS* dataset with handwriting text, end-to-end methods exhibit superiority over multi-stage methods due to their ability to jointly optimize text detection and recognition. 
We conduct further studies on this characteristic in the further discussion section and appendix.

\subsection{Ablation Study}
In the ablation experiments, we conduct a thorough analysis of the effectiveness of each of our contributions, including pre-training, prompt generation, and content-aware token merge module. 

\noindent\textbf{Impact of Pre-training.}
The role of pre-training in augmenting the performance of large-scale models is widely recognized in the research community. In this paper, we investigate the effect of three pre-training tasks on the model's performance in the context of the SROIE dataset. We present the experimental results in Table~\ref{table:abl_pre}, which demonstrate the efficacy of pre-training in enhancing the information extraction performance of the model.

\begin{table}[t] 
   \centering
   
   \begin{tabular}{cc}
   \toprule 
   method & F1 \\
   \midrule
   query to segmentation & 59.3 \\
   + text to segmentation   & 82.5\\
   + segmentation to text & 85.8\\
   \bottomrule
   \end{tabular}
   \caption{Effect of pre-training method on the SROIE dataset.}
   \label{table:abl_pre}
\end{table}

\noindent\textbf{Impact of Generation Manners.}
As an end-to-end generative model, SeRum enables complete extraction of essential information by generating a string sequence that includes all relevant keys, referred to as SeRum-total. 
To achieve this, we utilize a Donut-like method \cite{kim2022donut} to serialize a json style string, with each key-value pair represented as \textit{ \textless s\_key\textgreater value\textless e\_key\textgreater}, where both \textit{\textless s\_key\textgreater} and \textit{\textless e\_key\textgreater} are added to the tokenizer as special tokens. 
The complete json is encoded into a single sequence, which is then decoded to extract all the necessary information in one pass.

Table~\ref{table:results} displays the performance differences between SeRum-total and SeRum-prompt in the information extraction task. 
Our results indicate that the one-pass extraction method of SeRum-total shows varying degrees of degradation compared to the prompt-based approach, particularly on the CORD dataset. 
This degradation may be attributed to the fact that our pre-training is a pseudo-OCR task, limiting the model's ability to comprehend complex structures. 
Nonetheless, parsing the CORD dataset into a dictionary query leads to a significant enhancement of the model's performance. Moreover, the substantial improvement observed in the Ticket and SROIE datasets further validates the effectiveness of prompt generation.

\noindent\textbf{Impact of Content-aware Token Merge.}
Table~\ref{table:token_merge_ratio} presents the impact of the content-aware token merge mechanism on the recognition accuracy and decoding speed of SeRum.
We observe that with an increasing token keep ratio, the recognition accuracy of SeRum improves gradually, peaking at a merge ratio of 10\%.
However, surpassing this ratio causes a decline in recognition accuracy.
Regarding decoding speed, as the token keep ratio decreases, the decoding speed of SeRum improves gradually.
These findings suggest that the content-aware token merge mechanism in SeRum is capable of enhancing both recognition accuracy and decoding speed. Furthermore, the ratio can be dynamically adjusted to achieve varying speeds during inference.

\begin{table}[t] 
   \centering
   \begin{tabular}{cccc}
   \toprule 
   Token keep ratio $\alpha$ &  F1 & Text decoder latency(ms) \\
   \midrule
   2\% & 72.5 & 194 \\
   5\% & 83.2 & 198 \\
   10\% & 85.8 & 209 \\
   20\% & 84.8 & 225 \\
   30\% & 84.9 & 231 \\
   50\% & 84.9 & 234 \\
   100\% & 84.9 & 306 \\
   \bottomrule
   \end{tabular}
   \caption{Effect of token keep ratio on the SROIE dataset. The latency measures on a T4 GPU.}
   \label{table:token_merge_ratio}
\end{table}

\subsection{Further discussion}
\textbf{Case Study.} 
In order to further investigate the effectiveness and problems of the scheme, we further analyzed the differences with the SOTA method on multiple data sets, as well as the accuracy of the seleted region mask. 
As shown in Figure~\ref{fig:showcase}, the benefit of SeRum model is a higher accuracy of word recognition by focusing on important relevant details. 
Especially in some confusing scenes, it is often better to combine the contextual information of the text. See the appendix for more details and cases.

\begin{figure}[t]
   \centering
   \includegraphics[width=0.5\textwidth]{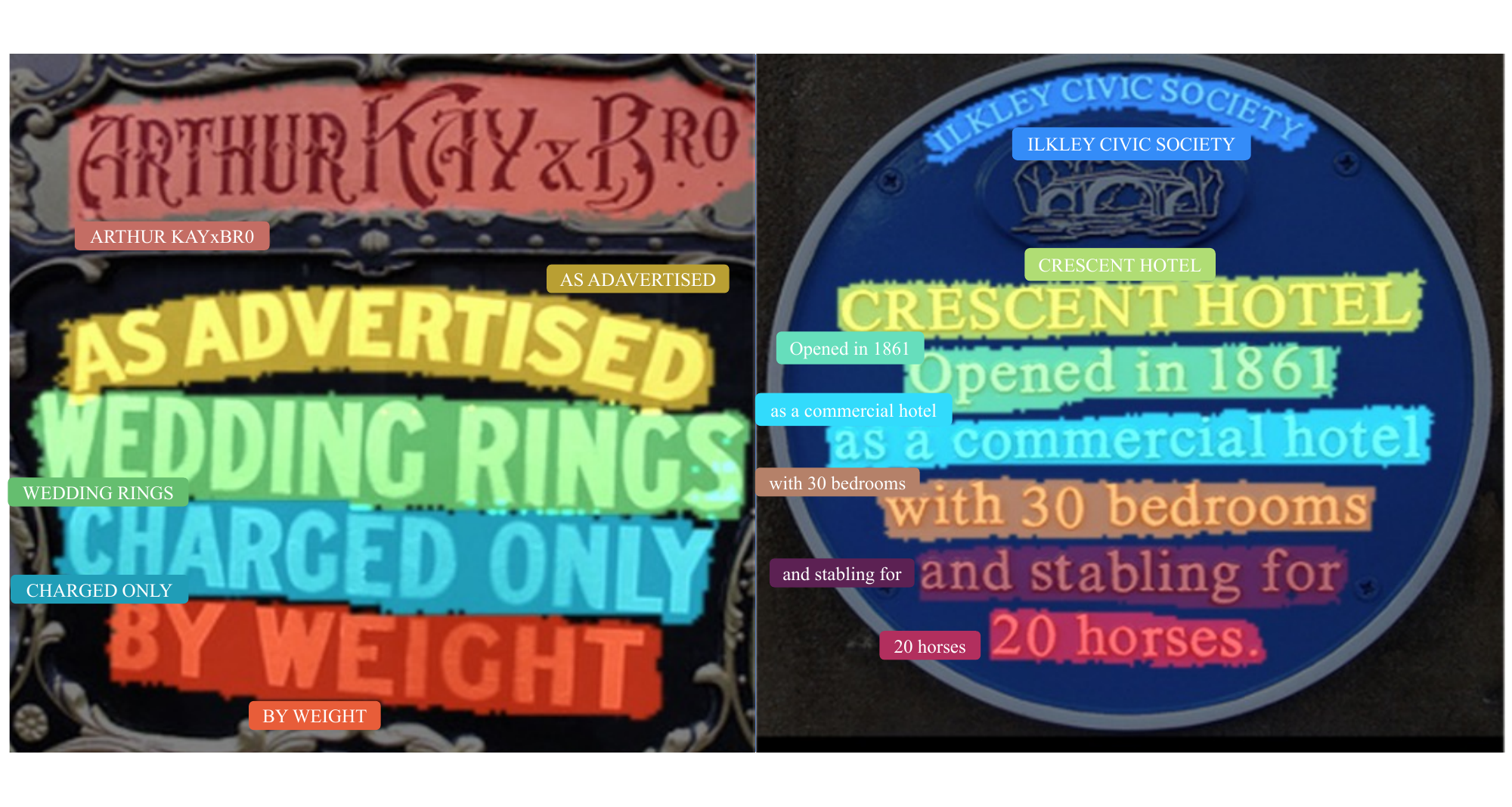}
   \caption{Results of text spotting. SeRum demonstrates significant advantages in handling documents with complex formats, including WordArt, curved text and \textit{etc.} Best viewed in color.} 
   \label{fig:spotting}
\end{figure}

\textbf{Generalization to Other Tasks.}
SeRum's ability to generalize to other tasks was assessed by testing its performance on text spotting, as illustrated in Figure~\ref{fig:spotting}.
Text spotting involves detecting and recognizing text in natural images, such as street views or signs. 
In our evaluation, we used the widely-used CTW-1500~\cite{ctw1500} dataset, and the experimental results are summarized in Table \ref{table:textspotting}.
The experimental results show that SeRum attains competitive performance across multiple tasks, emphasizing its efficacy in diverse applications. 
We suggest that this approach could be promising for various document understanding tasks.
\begin{table}[t] 
   \centering
   \begin{tabular}{cc}
   \toprule 
   method & F1  \\
   \midrule
   TextDragon~\cite{textdragon} & 39.7   \\
   ABCNet~\cite{abcnet}  & 45.2\\
   SPTS v2 ~\cite{liu2023spts} & \textbf{63.6} \\
   SeRum & 41.8\\
   \bottomrule
   \end{tabular}
   \caption{Performances on CTW-1500 text spotting datasets.}
   \label{table:textspotting}
\end{table}

\section{Conclusion}
In conclusion, we have presented a novel query-based approach for image-based document understanding that does not rely on OCR, but instead uses a visual encoder and a text decoder. 
Our approach encodes the prompt and extracts the relevant token regions through a context-aware token merge mechanism between the prompt and image features, 
and with a decoder to generate the local document understanding results.

Our method achieves competitive performance on several public datasets for document information extraction and text spotting, and is almost comparable to the two-stage solution while being more efficient and enabling parallel generation based on the prompt. The proposed model has the potential to be applied in real-world scenarios where traditional OCR-based approaches are not feasible or efficient.

\clearpage
{\small
\bibliographystyle{ieee_fullname}
\bibliography{egbib}
}
\clearpage
\appendix
\begin{center}
   \textbf{Appendix}
\end{center}

This supplementary material presents a comparative case study of SeRum with end-to-end methods and OCR-dependent methods for handling challenging images. 
The evaluation utilizes several test sets, such as SROIE~\cite{huang2019icdar2019}, CORD~\cite{park2019cord}, Ticket~\cite{guo2019eaten}, and DocVQA~\cite{docvqa}.

\textbf{Comparison results with end-to-end methods.}
SeRum is compared with Donut~\cite{kim2022donut}, the current state-of-the-art end-to-end document understanding method, which decoding text directly from image features. 
However, Donut suffers from generating overly long results, leading to instability, and attention mechanism deviation and confusion. 
In contrast, SeRum excels at decoding the form of localized visual tokens of interest, leading to significant improvements in both of these drawbacks.

\begin{figure}[t]
   \centering
   \includegraphics[width=0.5\textwidth]{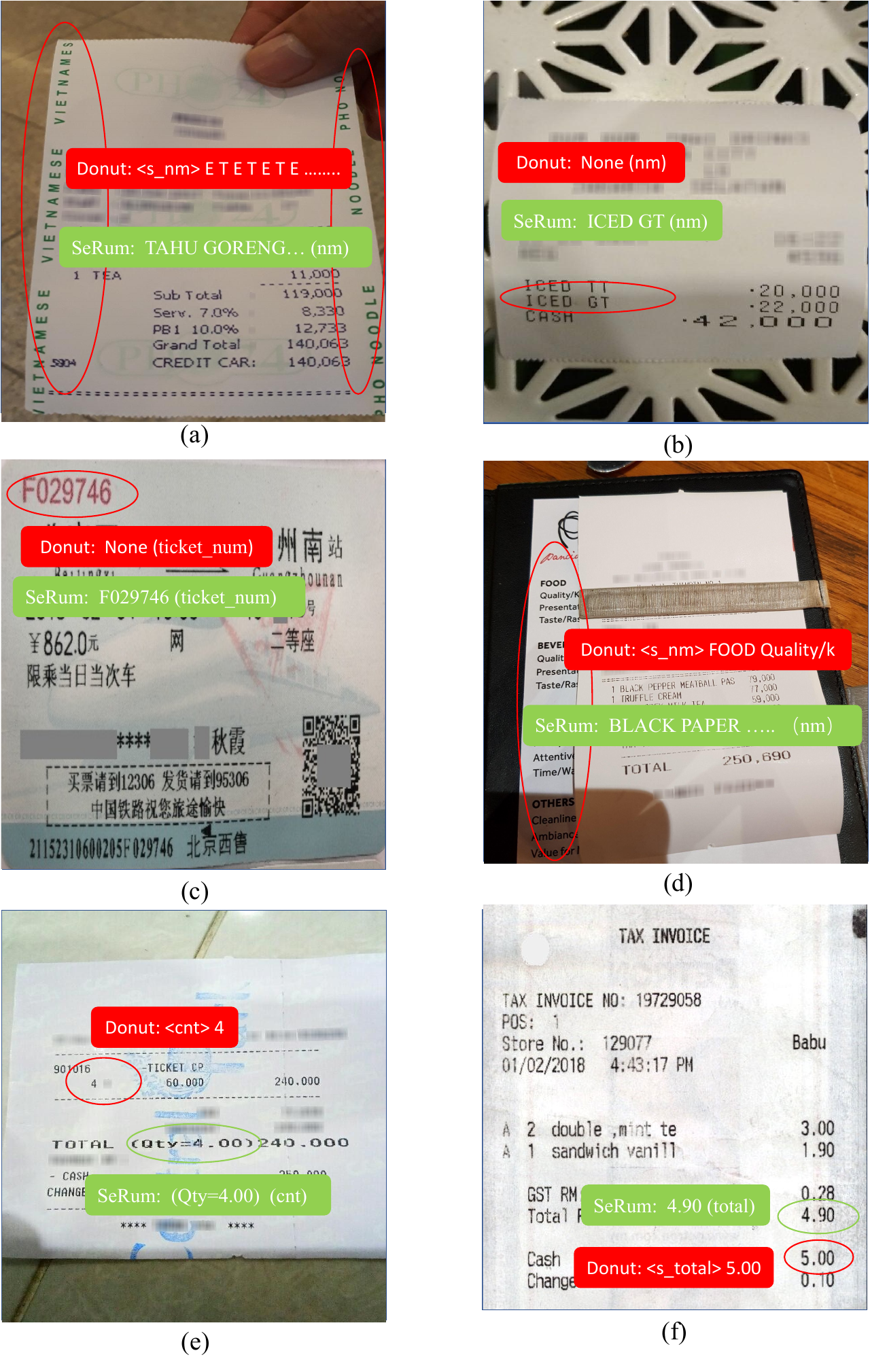}
   \caption{ Challenging cases encountered in our study, including instances of redundant text on the border, superimposed images, \textit{etc.} The red and green boxes represent the outputs of Donut and SeRum, respectively. Best viewed in color.}
   \label{fig:donut}

\end{figure}

As illustrated in Figure~\ref{fig:donut}.(a) to (c), Donut generates an abnormal sequence of text due to the interference of redundant characters, and it cannot correctly parse all the key information. 
In contrast, SeRum possesses the ability to identify the key area of interest and perform decoding process in isolation. 
Additionally, as shown in Figure~\ref{fig:donut}.(d) to (f), Donut exhibits a tendency to misinterpret the location of text, whereas SeRum is capable of correctly identifying the text and its location within the image. 
Overall, SeRum demonstrates a superior performance relative to Donut.

\begin{figure*}[t]
   \centering
   \includegraphics[width=\textwidth]{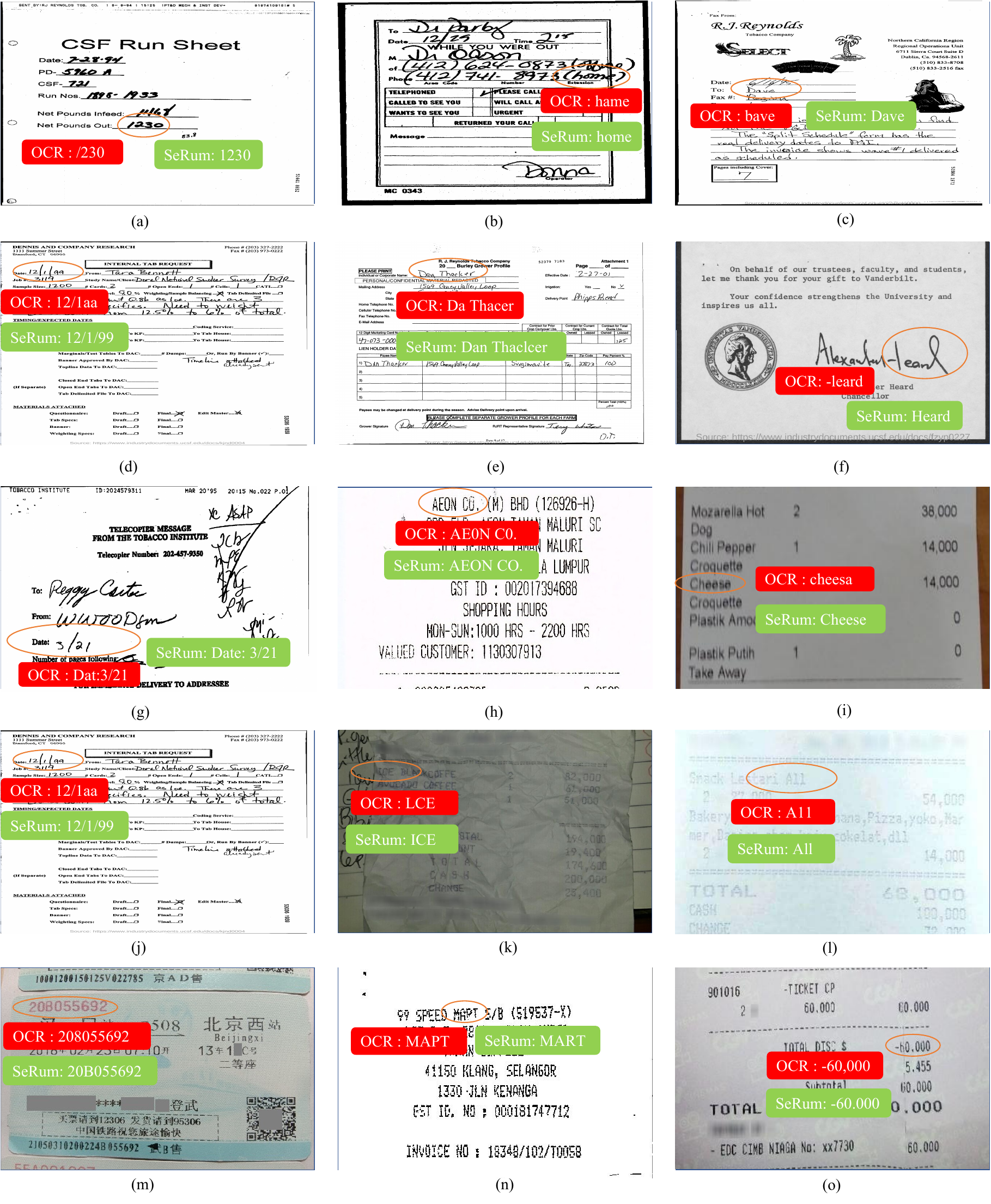}
   \caption{Examples of challenging cases, such as handwritten, blurred, and missing characters, illustrating the difficulties faced by OCR systems. 
   These characters are known to pose difficulties in accurate recognition.
   The red and green boxes represent the outputs of OCR engine and SeRum, respectively. Best viewed in color.}
   \label{fig:appendix_total}
\end{figure*}

\textbf{Comparison results with OCR-dependent methods.}
This section evaluates the performance of SeRum on handwritten or blurry text images. 
Handwritten text recognition poses a significant challenge to OCR systems due to the inherent complexity and variability of handwritten fonts. 
Handwritten characters exhibit a high degree of variation in shape, size, slant, \textit{etc.} as shown in Figure~\ref{fig:appendix_total}.(a) to (g). 
Besides, the stability of the system can also be significantly impacted by the presence of blurry text, as exemplified in Figure~\ref{fig:appendix_total}.(h) to (o).

The SeRum model simplifies the character recognition pipeline by integrating all stages into a single model, achieving end-to-end optimization that improves accuracy and reduces error propagation. 
Additionally, the model utilizes attention mechanisms to extract robust features from input images and effectively utilize contextual information. 

Our findings indicate that the SeRum method can synthesize the context and help improve the OCR recognition results.
For example, in Figure~\ref{fig:appendix_total}.(d), the SeRum model identifies the word `home' after a phone number, indicating that it is a home phone rather than a less common term like `hame'. As shown in 
Figure~\ref{fig:appendix_total}.(n), the SeRum model distinguished `MART' from `MAPT', despite the visual similarity of the latter due to vagueness.

\end{document}